\documentclass{article}

\usepackage{microtype}
\usepackage{graphicx}
\usepackage{booktabs} 

\usepackage{hyperref}

\usepackage[accepted]{icml2021}

\newcommand{\MSE}[1]{\texttt{MSE}}
\newcommand{\GEN}[1]{\texttt{GEN}}
\newcommand{\OJA}[1]{\texttt{OJA}}
\newcommand{\INEL}[1]{\texttt{INEL}}
\newcommand{\SPARSE}[1]{\texttt{SPARSE}}
\usepackage{subcaption}
\usepackage{color}
\usepackage{array,multirow}
\newcolumntype{H}{>{\setbox0=\hbox\bgroup}c<{\egroup}@{}}

\usepackage{tabularx}
\usepackage{adjustbox}
\usepackage{wrapfig}
\usepackage{enumitem,amssymb}
\newlist{todolist}{itemize}{2}
\setlist[todolist]{label=$\square$}
\usepackage{pifont}
\let\subparagraph\paragraph
\usepackage[compact]{titlesec}

\usepackage[switch]{lineno}

\usepackage{multibib}
\newcites{latex}{\LaTeX-Literature}

\icmltitlerunning{Neuromodulated Neural Architectures with Local Error Signals for Memory-Constrained Online Continual Learning}

\begin{document}

\twocolumn[

\icmltitle{Neuromodulated Neural Architectures with Local Error Signals for Memory-Constrained Online Continual Learning}




\begin{icmlauthorlist}
\icmlauthor{Sandeep Madireddy}{mcs}
\icmlauthor{Angel Yanguas-Gil}{amd}
\icmlauthor{Prasanna Balaprakash}{mcs}
\end{icmlauthorlist}

\icmlaffiliation{mcs}{Mathematics $\&$ Computer Science Division, Argonne National Laboratory, Lemont, IL, USA}
\icmlaffiliation{amd}{Applied Materials Division, Argonne National Laboratory, Lemont, IL, USA}

\icmlcorrespondingauthor{Sandeep Madireddy}{smadireddy@anl.gov}


\vskip 0.3in
]



\printAffiliationsAndNotice{}  

\begin{abstract}

The ability to learn continuously from an incoming data stream without catastrophic 
forgetting is critical for designing intelligent systems. Many existing approaches 
to continual learning rely on stochastic gradient descent and its variants. 
However, these algorithms have to implement various strategies, such as memory buffers 
or replay, to overcome well-known shortcomings of stochastic 
gradient descent methods in terms of stability, greed, and short-term memory.

To that end, we develop a biologically-inspired light weight neural network architecture 
that incorporates local learning and neuromodulation to enable input processing over data 
streams and online learning. Next, we address the challenge of hyperparameter selection for 
tasks that are not known in advance by implementing transfer metalearning: using a Bayesian 
optimization to explore a design space spanning multiple local learning rules and their hyperparameters, 
we identify high performing configurations in classical single task online learning and we transfer them to 
continual learning tasks with task-similarity considerations. 

We demonstrate the efficacy of our approach on both single task and continual learning setting. 
For the single task learning setting, we demonstrate superior performance over other 
local learning approaches on the
MNIST, Fashion MNIST, and CIFAR-10 datasets. Using high performing configurations 
metalearned in the single task learning setting, we achieve superior  
continual learning performance on Split-MNIST, and Split-CIFAR-10 data as compared with other 
memory-constrained learning approaches, and match that of the state-of-the-art memory-intensive 
replay-based approaches. 

\end{abstract}

\section{Introduction}

Online continual learning
addresses the scenario
where a system has to learn and process
data that are streamed continuously,
often without restrictions in terms of the distribution of data within
and across tasks and without clearly identified task boundaries. These
conditions are present in many
real-life applications. Online continual learning algorithms aim to mitigate catastrophic forgetting at both the data-instance and task level~\cite{chen2020mitigating}.
However, in some cases, such as on-chip AI at the edge, additional considerations such as resource limitations in the hardware, 
data privacy, or data security, threaten the ability to implement these algorithms at scale.

The challenges in the online continual learning scenario are therefore two-fold: first, resource constraints impose limitations to algorithms, the extreme case being learning just from streaming data with no control of when and how this data is presented. Second, systems need to learn new tasks from data that may have not been seen before deployment. These challenges run counter to the optimal conditions required for optimization using stochastic gradient descent.

In this work, we explore how to overcome the inherent
shortcomings of stochastic gradient descent methods for online and continual learning, building on some fundamental hypotheses extracted from
biological systems, where continual learning is ubiquitous, to design memory-constrained online learning architectures.

In particular, we look beyond replay to explore 
the following hypotheses:
\begin{itemize}[noitemsep]
    \item \emph{Heterogeneous plasticity is a key factor promoting continual learning in neural networks} In biological networks, synaptic plasticity is typically confined to a subset of the network. This requires establishing robust priors that can provide strong representations and enhance the depth of the network.
    \item \emph{Local synaptic plasticity rules can implement adaptation and consolidation to mitigate catastrophic forgetting} Modulated learning in biological systems is driven by local synaptic plasticity rules that can be
    mathematically expressed as:
    \begin{equation}
        W(t+1) = f\left(\mathbf{x}_e, \mathbf{x}_o,
        \mathbf{x}_m; W_{t}, \beta_k\right)
    \end{equation}
    Where $W$ are the synaptic weights, $\mathbf{x}_e$,
    $\mathbf{x}_o$, $\mathbf{x}_m$, are pre, post-synaptic, and modulatory inputs, and $\beta_k$
    are a set of hyperparameters.
    The set formed by a feedforward network and local learning rules can be viewed as a recurrent neural network that is Turing complete without the need
    to access external memory. This potentially allows the implementation of rules that create long-term effects in plasticity akin to synaptic consolidation mechanisms.
    
    \item \emph{Transfer metalearning can help optimize the online learning capabilities for previously unseen tasks} Optimal hyperparameters depend on the type of input, network architecture, and learning modality.
    Here, we explore the transferability of hyperparameters to previously unseen tasks and data, and learning modalities. 
\end{itemize}

To explore these hypotheses, we develop a biologically-inspired network architecture to carry out online supervised learning with input streams under single task, task-incremental, and class-incremental learning. Our approach does not require an 
explicit memory buffer and employs light weight neutral network, hence effectively addressing the 
resource constraint challenge in online continual learning scenarios.

We then use this architecture to explore transfer metalearning: using a Bayesian optimization to explore a design space spanning 
multiple local learning rules and their hyperparameters, we identify high performing configurations 
in classical single task learning and study their transfer to continual learning tasks with task-similarity considerations.

We demonstrate accuracies in the single task learning scenarios (with the MNIST, Fashion MNIST and CIFAR-10 datasets) 
that are on-par or outperform other local learning approaches employing shallow neural networks, 
all with a fraction of the model parameters and runtime epochs. We then transfer metalearn the 
configurations to an online task-incremental learning scenario
(with Split-MNIST and Split-CIFAR-10 datasets) and demonstrate accuracies that outperform the memory-free approaches and match the approaches that employ memory buffers, with a shallow network and without memory replay. We then analyze the transfer to a more challenging class-incremental learning setting and provide design principles that enabled us to obtain accuracy that is significantly better than other memory-free approaches and is comparable to the approaches that employ the memory buffers. Finally, we analyze the transfer metalearning across datasets (on single task learning) and across tasks (in the continual learning) and its correlation to transfer coefficients based on dataset similarities.

\section{Related Work}

Several incremental learning approaches have been presented in the literature, which can loosely be categorized into three classes: (1) novel neural architectures or customization of the common ones; (2) regularization strategies that impose constraints to boost knowledge retainment; and (3) metalearning, which uses a series of tasks to  learn a common parameter configuration that is easily adaptable for new tasks.

Algorithms in the first category include bio-inspired dual-memory architecture~\cite{parisi2019continual}; progressive neural networks~\cite{rusu2016progressive} that explicitly support information transfer across sequences of tasks through network expansion; and deep generative replay~\cite{shin2017continual}, which proposed a cooperative dual model architecture framework, inspired by hippocampus, that retains past knowledge
by the concurrent replay of generated pseudo data. Gradient episodic memory~\cite{lopez2017gradient} and generative replay with feedback connections~\cite{van2018generative} are other examples of replay-based approaches in this category. The replay-based approaches maintain an exemplar data buffer of samples from previous tasks. Trade-offs have to be made on the size of this buffer and the bias introduced by the use of data from just the most recent task. 

The second category consists of algorithms such as elastic weight consolidation (EWC)~\cite{kirkpatrick2017overcoming} that computes synaptic importance using a Fisher importance matrix-based regularization; synaptic intelligence (SI)~\cite{zenke2017continual}, whose regularization penalty is similar to EWC but is computed online at per-synapse level;  and learning without forgetting~\cite{li2017learning} applies a distillation loss on the attention-enabled deep networks seeking to minimize task overlap.

The third category consists of metalearning-based approaches that consist of an inner loop that learns parameters in a prediction network and the outer loop that learns the parameters in a representation learning network. These loops are updated using different strategies and data splits at meta-training and meta-testing. Recent approaches in this category include online metalearning (OML)~\cite{javed2019meta}, neuromodulated metalearning algorithm (ANML) ~\cite{beaulieu2020learning}, and incremental task-agnostic metalearning (iTAML)~\cite{rajasegaran2020itaml}. In terms of memory, a naive metalearning approach will maintain full task history to update the parameters for each traversal through the loop when new task is observed. OML, ANML algorithms keep all the tasks in memory but randomly choose the task sequences for meta-training (not just from task history) and meta-testing updates. iTAML on the other hand uses data from the current task and a exemplar data (memory buffer) from the previously seen tasks to metalearn. Due to this memory and computation overhead, these approaches are not particularly suited for online continual learning. Hence, we do not include a comparison with this class of approaches.

Most of these approaches however, have been designed for the traditional continual learning scenarios where non-stationarity is assumed between tasks while the data within tasks are assumed to be i.i.d and multiple passes are made through them. Recent works such as~\cite{aljundi2019gradient,aljundi2019online,chaudhry2018efficient,chen2020mitigating}, started to look at the more challenging online continual learning scenario where the data is seen only once, which resembles real-world continual learning. In this scenario, however, majority of the works~\cite{chen2020mitigating} have looked at task-incremental learning setting where there the task labels are explicitly provided at test time. Only a few works~\cite{aljundi2019gradient,aljundi2019online,Lee2020A} have employed the class-incremental learning scenario which doesn't have access to task labels, and hence is much harder. The restriction of learning with only a single pass over the data makes this even harder. To handle this scenario, most of the approaches resort to memory buffers~\cite{hsu2018re,buzzega2020dark} and auxiliary data~\cite{zhang2020class}. To this end, we propose a memory-free online continual learning approach that incorporates synaptic plasticity-based local learning and neuromodulation into light weight neural network that mitigates the need for explicit memory buffer, while providing good accuracy in both task and class-incremental learning scenarios.

\section{Online learning architecture}

In this work we focus on the following problem:
given a stream of inputs $\mathbf{u}(t)$ and potentially
sparse modulatory signals $\mathbf{x}_m(t)$ conveying
label information, we seek to develop a system that
optimizes the ability to learn and process inputs
so that:
\begin{eqnarray}
\mathbf{x}_0(t) & = &  F\left(\mathbf{u}(t),\mathbf{x}_m(t);
S(t)\right)\\
S(t+1) & = & F_s\left(S(t),\mathbf{u}(t),\mathbf{x}_m(t),
\mathbf{x}_0(t)\right)
\end{eqnarray}
where $S(t)$ represents the internal state of the
system.

Our approach is to decompose this problem into four different coupled systems:
\begin{eqnarray}
\mathbf{x}_e & = & F_e(\mathbf{u}, W_e) \\
\mathbf{x}_o & = & F_l(\mathbf{x}_e, W_l) \\
W(t+1) & = &  f\left(\mathbf{x}_e, \mathbf{x}_o,
        \mathbf{x}_m; W(t), \beta_k(t)\right) \\
\beta(t+1) & = &  f_\beta\left(\mathbf{x}_e, \mathbf{x}_o,
        \mathbf{x}_m; W(t), \beta_k(t)\right)
\end{eqnarray}
subject to the initial conditions $W(0)$, $\beta(0)$. 
The first two equations represent the
feedforward inference system, broken into a feature extraction
and a learning modules, and the last two equations
represent the synaptic plasticity mechanism and
the evolution of the state of the network, represented
by the evolution of hyperparameters $\beta$. 
Since our system does not store batches of prior
data or configurations, any memory of prior
processes has to be built into
the equations for $W$ and $\beta$. 
In contrast to conventional machine learning approaches,
there is no separation between architecture and
learning algorithm: the ability to learn is defined
by the choice of $f$, $f_\beta$, and their hyperparameters.

\begin{figure}[!t]
\centering
  \includegraphics[width=\linewidth]{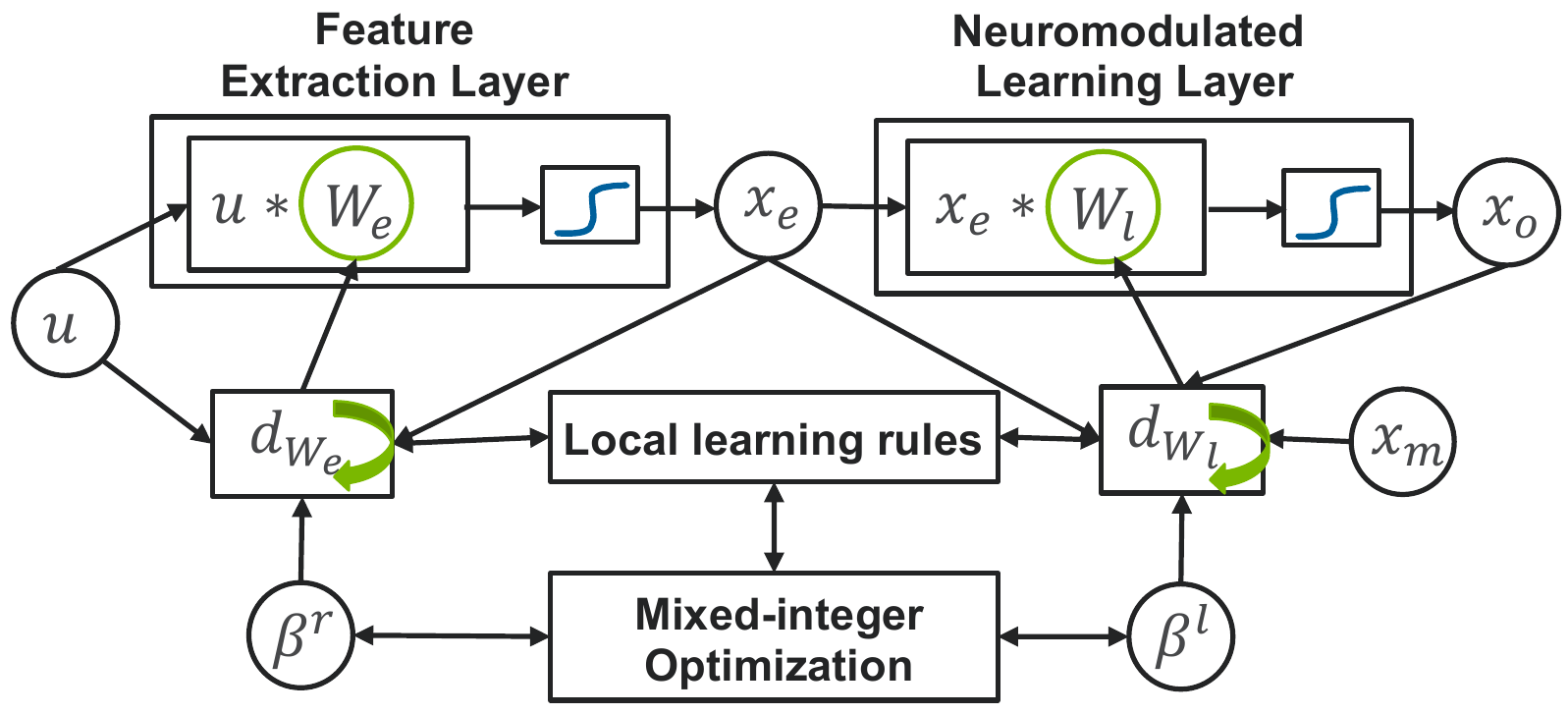}
\caption{Multilayer neuromodulated architecture that consists of feature extraction and neuromodulated learning layers that incorporate synaptic plasticity mechanisms through local learning rules.} 
\label{fig:architecture}
\end{figure}

Our approach allows us to define a configuration
space
$\left\{F_e, F_l, f, f_\beta, W(0), \beta(0)
\right\}$ that can be searched for optimal configurations. We describe the specific
components chosen in this work in the next sections. A typical
configuration is shown in Figure \ref{fig:architecture}

\subsection{Feature extraction layer} 

For single channel datasets we have considered a sparse layer that follows
the same design pattern found in the cerebellum
and the insect's mushroom body.\cite{Litwin_2017,ayg_memristors}
A $N$-dimensional input is projected into a much larger $M$-dimensional space, so that each
neuron receives $K$ inputs, with $K \ll N$.
We use dynamic thresholding in the sparse layer
followed by a rectified linear unit to ensure
that only the most salient features are included in the representation:
\begin{equation}
\label{eq:dyn_thresh}
    \mathbf{x_e} = \mathrm{ReLU}\left(\mathrm{W_e} \mathbf{u} - \mu - \beta_r^1 \sigma\right),
\end{equation}
where $\mathrm{W_e}$ is the sparse projection matrix,
$\mu$ and $\sigma$ are the mean and standard deviation of the
matrix multiplication product
$\mathrm{W_e} \mathbf{u}$, and $\beta_r^1$ is a constant controlling the cutoff.
This approach highlights correlations between different input channels and represents the broadest possible prior for feature extraction, since it does not assume any spatial dependence or correlations between inputs. $\mathrm{W_e}$, $K$, and $\beta_r^1$ 
are all hyperparameters that can be adjusted to the network.

For experiments involving 3-channel images, we transfer feature extractors from various pre-trained models. Here we have used the wide-ResNet50~\cite{zagoruyko2016wide} model pre-trained on ImageNet data~\cite{deng2009imagenet} after experimenting with a wide variety of pre-trained model and data combinations.

\subsection{Neuromodulated learning layer} 

This
comprises one or more all-to-all 
connected layers where online supervised learning occurs. Synaptic weights are treated as first-class network elements that can be updated in real time through a series of local learning rules codified in the learning component.

\subsection{Learning Component}

It comprises the recurrent version of the network, implementing local learning rules. In this work we 
have focused on modulatory learning rules, which involve a modulatory signal controlling synaptic plasticity. Here, we consider the following synaptic plasticity rules.

\begin{itemize}[leftmargin=*]
    \item {\it Generalized Hebbian model (GEN):} the generalized 
    Hebbian model is a modulated version of the covariance
    rule commonly used in neuroscience. The synaptic
    weight evolution is given by
    \begin{equation}
        \Delta W_l = l_r\mathbf{x}_m(\mathbf{x}_e - \beta_1^l)
        (\mathbf{x}_o - \beta_2^l).
    \end{equation}
    It has long been known that this rule is unstable 
    and that clamping of a regularization mechanism
    needs to be included in order to keep the synaptic weights
    bounded. This rule is at the core of some of
    the neuromodulation-based approaches recently developed \cite{Miconi2018}.

    \item {\it Oja's rule (OJA):} a modification of the basic Hebb's rule
    providing a normalization mechanisms through a first-order loss term \cite{Oja1982}:
    \begin{equation}
        \Delta W_l = l_r 
        \mathbf{x}_m(\mathbf{x}_e\mathbf{x}_o - 
        \beta_1^l x_o^2 W_l).
    \end{equation}
    
    \item {\it MSE, non-Hebbian  rule:} It is based on recent experimental results on synaptic
    plasticity mechanisms in the mushroom body  \cite{Hige_nonhebbian_2015}. The key
    assumptions are that learning is independent of postsynaptic activity and that instead postsynaptic activity modulates
    regulates synaptic plasticity:
     \begin{equation}
        \Delta W_l = l_r
        (\mathbf{x}_m-\mathbf{x}_o)\mathbf{x}_e .
    \end{equation}
    This rule is consistent with the MSE cost function used
    in stochastic gradient descent methods. 
    
    \item {\it Modulated Inelastic rule:} This is a novel rule introduced in this work that
    considers
    a synapse-specific window for plasticity based on how significant a weight $W_{ij}$ is with respect to
    all presynaptic weights of the same neuron $W_{ij}$, so that:
    \begin{equation}
        l_{ij} = H\left(1-\beta |W_{ij}-\mu|\right)
    \end{equation}
    Where $H(\dot)$ is the Heaviside
    function and $\mu$ is the average
    synaptic weight of each neuron.
    The resulting $l_{ij}$ is then fed to an MSE rule.
    This rule introduces a simple mechanism for memory consolidation. Note, though, that this consolidation is not irreversible: the evolution of the remaining weights can bring weights from frozen to active by reducing its significance.
    
\end{itemize}

\noindent We have treated these models as categorical variables that can be swapped during the metalearning optimization process. Moreover, we have added two conditions that allow us to externally tune the learning rate as a function of time and to provide symmetric
or positive clamping of the synaptic weights in order to
prevent instabilities in the algorithms.

\subsection{Mixed-Integer Black-box Optimization Framework}

The parameter space in the proposed multilayer neuromodulated architecture is composed of categorical variables (e.g., the selection of the local learning rule), integer parameters (e.g., the dimension of hidden layer), and continuous parameters in each of the learning rules. We adopt a parallel asynchronous-model-based search approach (AMBS)~\cite{Balaprakash_DH_2018} to find the high-performing parameter configuration in this mixed (categorical, continuous, integer) search space. Our AMBS approach consists of sampling a number of parameter configurations and progressively fitting a surrogate model over the parameter configurations' accuracy metric space. This surrogate model is asynchronously updated as new configurations are being evaluated by the parallel processes, which are then used to obtain configurations that will be  evaluated in the next iteration.

Crucial to the optimization approach is the choice of the surrogate model, since this model generates the configuration to evaluate in the mixed search space. The AMBS adopts {\it random forest} approach to build efficient regression models on this search space. The random forest is an ensemble learning approach that builds multiple decision trees and uses bootstrap aggregation (or bagging) to combine them to produce a model with better predictive accuracy and lower variance. Another key choice for the AMBS approach is the acquisition function, which encapsulates criteria to choose the most promising configurations to evaluate next. Hence, the acquisition function is key to maintaining the  exploration-exploitation balance during the search. The AMBS adopts the {\it lower confidence bound} (LCB) acquisition function. We set the kappa to a large value, which increases exploration. We do so to accommodate the high variability we observed in the accuracy metrics within and across the local learning rules that led them to local minimum when a smaller value of kappa was used.

\section{Online Single Task and Continual Learning Experiments}

We have considered experiments where the the system
is subject to a stream of data and labels, evolving its internal configuration during a predetermined number of epochs. 
This constitutes a single episode. 
At the end
of the episode, the system is evaluated against the testing dataset to validate 
its accuracy. By concatenating multiple episodes involving different tasks and 
datasets, we can create a curriculum to evaluate the system's ability to carry 
out continual learning.

We consider three different learning modalities:

\noindent \textbf{(a) Single Task Online Learning:} There is a single episode 
in the curriculum, which stream all the data for the system to perform a single
task (e.g., multiclass classification);

\noindent \textbf{(b) Task-Incremental Continual Learning:} The system
is subjected to a sequence of
episodes, each comprising a specific 
tasks (e.g., two-class classification).
Each episode has a different
output head (e.g., two-class classifier).
Task labels are used both at
train and test time.

\noindent \textbf{(c) Class-Incremental Continual Learning:}
The episodes and the curriculum is similar to task-incremental learning except 
that the model learns with a single (shared) output head. The final accuracy measures the model's ability 
to predict on test data from all the classes, spanning across episodes. The learning algorithm utilizes the task labels only at train time, thus
this is much harder scenario and closer to the ``task-free" continual learning.

In order to search for the optimal
configuration, we use the testing
accuracy as the metric to guide
 the exploration of the configuration
 space.
During the optimization process, multiple episodes 
are run for systems with different architectures and specific hyperparameters. 
In all these cases, the system starts from identical starting conditions, so 
that no knowledge is transferred between episodes that do not belong to 
the same curriculum.
High performing configurations are then used to explore
the transferability of optimal learning conditions to other datasets and to different 
types of tasks.

\section{Results and Discussion}

\subsection{Single Task Online Learning}
In this experiment, we demonstrate the learning capability of our multilayer  neuromodulated learning framework on a single episode curriculum learning task, specifically multi-class classification.
We consider MNIST, Fashion MNIST~\cite{xiao2017fashion} (F-MNIST), and Extended MNIST~\cite{cohen2017emnist} (E-MNIST) and CIFAR-10
datasets because of the existence of benchmarks in continual learning. 
For each dataset, we jointly optimize over the local learning rules and their parameters
to find the optimal configuration.
For each configuration evaluated during the optimization, the model is run for $0.5$ epochs. The best accuracy obtained and the corresponding optimal parameters for all the datasets are shown in Table~\ref{tab:Ninc-acc}. 
In all cases, MSE and INEL, which is a modification of MSE with memory consolidation, lead to the highest accuracies.

\begin{figure}[ht!]
  \centering

    \includegraphics[width=230pt]{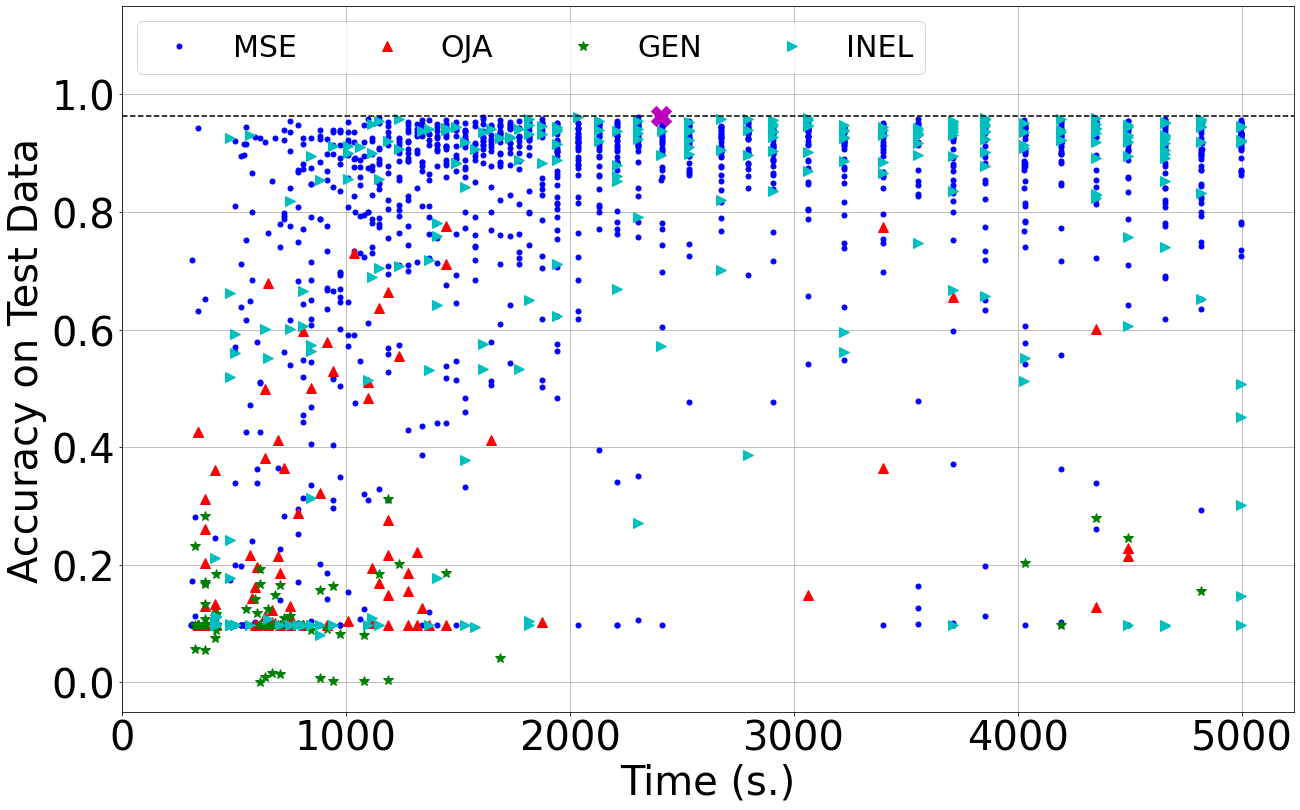} 
  \caption{Search trajectory obtained for MNIST dataset from the mixed-integer black-box optimization. The configurations are colored by their evaluated learning rule, with blue, red, green and cyan corresponding to \MSE., \OJA., \GEN., and \INEL. respectively. The accuracy on test data is plotted as a function of wallclock time, which highlights the diversity in the configurations evaluated as the search progressed.}
  \label{fig:search_traj}
\end{figure}

The search trajectory showing the learning rule and test accuracy of the configurations evaluated as a function of time is shown Fig.~\ref{fig:search_traj} for MNIST,
show how the initial configurations range across the different learning rules, but the algorithm quickly finds the potential learning rule and evaluates more configurations from it. 

The corresponding accuracies for each of the four datasets obtained using the optimal configurations, but run during $4$ epochs is shown in Table~\ref{tab:Ninc-acc4}.
We compare our approach with the kernelized information bottleneck (KIB)~\cite{pogodin2020kernelized} approach, which is also an SGD-alternative, local learning approach that is biologically plausible. In spite of having a shallow network (approximately 100k learnable weights) and operating in online learning with $1$ epochs of training, we obtain a test accuracy of $96.81$ on the MNIST data, $85.22$ on the F-MNIST data and $73.24$ on the CIFAR-10 data. This is very close to the accuracy of $98.1$ obtain by KIB (trained with 3-layer, each with 1024 neurons of fully connected network for 100 epochs) on the MNIST data while significantly outperforming KIB on CIFAR-10 data. Our accuracies are on par with (or outperform) other SGD-based shallow network architectures~\cite{YanguasGil_MSE_2019,xiao2017_online,cohen2017emnist}. 

\begin{table*}[t]
\caption{Optimal configurations obtained for MNIST, F-MNIST, E-MNIST, and CIFAR-10 datasets using the mixed-integer black-box optimization approach, where each configuration was evaluated after $0.5$ epochs of training.}
\begin{center}
\begin{adjustbox}{max width=\textwidth}
\begin{tabular}{|c|c|c|c|c|c|c|c|c|c|c|c|c|c|}
    \hline
    Dataset & Acc. &  Srule & $N_H^{F}$ & $\beta_1^{F}$& $\beta_2^{F}$ & $\beta_3^{F}$ & $\gamma^{F}$ &  $\beta_1^{M}$ & $\beta_2^{M}$ & $\beta_3^{M}$ & $\gamma^{M}$ & $\alpha^{M}$  \\
    \hline
    \hline
    MNIST & $96.40$ & INEL & $11000$ & $0.040$ & $0.567$ & $1.257$ & -- &  -- & $0.045$ & -- & $0.153$ & --\\
    \hline
    MNIST & $96.16$ & MSE & $11000$ & $0.023$ & $0.429$ & $1.450$ & -- &  -- & -- & $0.860$ & $0.582$ & $0.011$ \\
    \hline
    F-MNIST & $85.13$ & MSE & $7000$ & $0.016$ & $0.773$ & $0.775$ & -- &  -- & -- & $0.671$ & $0.145$& $0.287$\\
    \hline
    E-MNIST & $96.16$ & MSE & $9000$ & $0.027$ & $0.306$ & $1.918$ & -- & -- & -- & $0.531$ & $0.189$& $0.350$\\
    \hline
    CIFAR-10 & $77.96$ & INEL & $11000$ & $0.004$ & $0.826$ & $0.672$ & --  & -- & $0.021$ & -- & $0.601$ & --  \\
    \hline
    CIFAR-10 & $77.37$ & MSE & $11000$ & $0.007$ & $0.857$ & $1.012$ & --  & -- & -- & $0.737$ & $0.481$ & $0.284$  \\
    \hline

\end{tabular}
\end{adjustbox}
\end{center}
\label{tab:Ninc-acc}
\end{table*}

\begin{table}[ht]
\caption{Classification accuracy in the single task online learning scenario for MNIST, F-MNIST, and CIFAR-10 datasets after four epochs using the optimal configurations learned through the optimization framework.}
\begin{center}
\begin{adjustbox}{max width=0.45\textwidth}
\begin{tabular}{|c|c|c H|c|}
    \hline
    
     Algo/Data & MNIST & F-MNIST & E-MNIST & CIFAR-10\\
    \hline
    pHSIC (Shallow), 100 epoch & 98.1 &  88.8 & - & 46.4 \\
    \hline
     Ours, 1 epoch & 96.81 & 85.22 & 97.36 & 73.24\\
    \hline
\end{tabular}
\end{adjustbox}
\end{center}
\label{tab:Ninc-acc4}
\end{table}

\subsection{Continual Learning}
\subsubsection{Task-Incremental Learning}
In this experiment, we demonstrate the learning capability of our approach in the task-incremental learning scenario
using the
Split-MNIST~\cite{farquhar2018towards} and the Split-CIFAR-10 continual learning benchmarks that have been extensively adopted in the literature~\cite{shin2017continual,zenke2017continual,nguyen2017variational}. The Split-MNIST data is prepared by splitting the original MNIST dataset (both training and testing splits) consisting of ten digits into five two-class classification tasks. This defines a continual learning scenario in which the model sees these five tasks incrementally one after the other. The Split-CIFAR-10 is analogous to Split-MNIST.

In this scenario, we transferred the configuration for local learning obtained in the single task learning case. We used the configuration of MNIST for the Split-MNIST experiment and that of CIFAR-10 for the Split-CIFAR-10 experiment.  We then compare the accuracy of our approach against baselines, memory-free and memory buffer-based continual learning approaches, that are evaluated on the the online continual setting. As baselines, we consider \textbf{iid offline}, where the model is evaluated in a non-continual learning scenario and trained with multiple pass through the data, that are sampled iid. We also compare against \textbf{Fine-tuning}, where the model is trained continuously upon the arrival of new tasks without any strategies to avoid catastrophic forgetting. Among the memory-free approaches, we choose \textbf{Online EWC}~\cite{schwarz2018progress}, \textbf{SI}~\cite{zenke2017continual},  \textbf{LwF}~\cite{li2017learning}, and among the memory-based methods we compare with \textbf{A-GEM}~\cite{chaudhry2018efficient}, \textbf{iCaRL}~\cite{rebuffi2017icarl}, \textbf{GSS}~\cite{aljundi2019gradient}, \textbf{RPSNet}~\cite{NEURIPS2019_83da7c53}, \textbf{InstAParam}~\cite{InstaParam}, and \textbf{DER++}~\cite{buzzega2020dark} to cover the most recent works. All the approaches with Split-MNIST use a fully connected neural network with two layers, each with 400 nodes that are trained with a batch size of 10 for 1 epoch through each task. For Split-CIFAR-10, we follow other works~\cite{buzzega2020dark,aljundi2019online} and use the ResNet-18, which is trained with a batch size of 10 for 1 epoch per task. 

The comparison is shown in Table~\ref{tab:TaskINC-offline}, where we find that our approach obtains an accuracy of $99.60$ after transfer learning the configuration (with MSE rule) from single task learning on MNIST data. Similar transfer from the CIFAR-10 to Split-CIFAR-10 evaluation gives an accuracy of $94.65$. Both these experiments show that our approach outperforms the memory-free and memory-based approaches in the task-free continual learning scenario with a shallow network and without any explicit memory replay.

\begin{table*}[ht]
\caption{Classification accuracy for the task-incremental and class-incremental learning experiments on Split-MNIST and Split CIFAR-10 datasets. We employ three configurations of our approach, the configurations corresponding to that of INEL rule, MSE rule learned in the single task online learning. The third configuration is explicitly learned through optimization in the class-incremental learning scenario. } 
\begin{center}
\begin{adjustbox}{max width=0.9\linewidth}
\begin{tabular}{|c|c|c|c|c|c|}
 \hline
      &  &\multicolumn{2}{c|}{Task-Incremental Learning} &\multicolumn{2}{c|}{Class-Incremental Learning}\\
    \hline
          & Method & Split-MNIST & Split-CIFAR-10 & Split-MNIST & Split-CIFAR-10\\ 
    \hline
    \hline
    \multirow{2}{*}{Baseline}& iid-offline  & 99.50 & 95.43 & 95.93 &  80.64 \\
    & Fine-Tune  &  97.78 & 70.28 & 19.65 & 17.34   \\
    \hline
    \hline
    \multirow{3}{*}{\shortstack[l]{Continual Learning \\ Memory-free }}& Online EWC  &97.94 & 60.64 & 19.66 & 17.41 \\
     &SI&  97.84  & 60.20 & 19.81 & 17.67\\
     &LwF&  99.16 & 59.95 & 21.37 & 18.68 \\
     \cline{1-6}
     \multirow{8}{*}{\shortstack[l]{Continual Learning \\ Memory-based }}&A-GEM  &  99.31 & 68.82 & 50.36 & 17.94 \\
     &iCaRL  &  98.50  & 82.44 & 72.49 & 38.92\\
     &GSS  &  98.46  & 86.22 &  53.69 & 48.37 \\
     &RPSNet & -- & 67.0 & -- & -- \\
     &InstAParam  & -- & 83.8 & -- & -- \\
     & ER-MIR  & -- & -- & 87.6 & 40.0\\
     & CN-DPM & -- & -- & {\bf 93.81} & 47.11 \\
     &DER++  &  99.36 & 87.11 & 92.34 & 54.08 \\
     \cline{1-6}
     & Ours (transfer w/ INEL) &  65.25  & 81.20 & 21.56 & 22.33 \\
     & Ours (transfer w/ MSE ) &  {\bf 99.60} & {\bf 94.65} & 21.84 & 22.85 \\
     & Ours (Opt) &  --  & --  & 78.76 & {\bf 55.74} \\
    \hline
\end{tabular}
\end{adjustbox}
\end{center}
\label{tab:TaskINC-offline}
\end{table*}

\subsubsection{Class-Incremental Learning}

In the experiment, we demonstrate our approach on the more challenging class-incremental learning scenario
on the Split-MNIST and Split-CIFAR-10 datasets. We first transfer the best-performing parameter configuration 
learned in the single task learning case (similar to the Task-incremental scenario)(Table~\ref{tab:Ninc-acc}) 
and then employ the configuration obtained by explicitly optimizing the learning rule configuration for the 
class-incremental learning. We then study the impact of this on model accuracy and extract further insights 
on the performance differences. This transfer of parameter configuration from the single task learning to the 
continual learning scenario is further discussed in the next section.

In Table~\ref{tab:TaskINC-offline} we compare the accuracy of our approach in the class-incremental learning setting against baselines, memory-free and memory buffer-based continual learning approaches. We use the same baselines and memory-free approaches as those adopted in task-incremental case, but for the memory buffer-based, we also compare against \textbf{ER-MIR}~\cite{aljundi2019online} and \textbf{CN-DPM}~\cite{Lee2020A}, which are the most recently proposed approaches. The hyperparameters used for this case are same as that adopted in the task-incremental learning case.

The results obtained with the configuration transferred from the single task
case are higher than other memory-free algorithms, but still substantially lower than those achievable using memory-based methods. In contrast, when we optimize our architecture against the class-incremental learning problem itself, we see a significant increase of performance in both cases. In both tasks, the Inelastic rule leads to accuracies that are three times higher than the memory-free algorithms. In the case of CIFAR-10, the resulting performance is also on par with that of the memory-based DER++ algorithm. The higher performance of the Modulated Inelastic rule can be attributed to the stabilization of the most significant weights for each class. However, as shown in Eq. 12, this rule provides a relative stabilization and it is sensitive to changes in other synapses reaching the same neurons. This sensitivity is modulated by a hyperparameter in Eq. 12. Consequently, if weight consolidation is not critical for the single task learning, it is likely that the optimal configurations do not fully exploit this feature, leading to threshold values that are not compatible with class-incremental experiments, which require stabilization over longer periods of time.

on-memory-based incremental learning approaches considered. In addition, we obtain accuracy comparable to the state-of-the-art memory-based models for Split-MNIST data and outperform all the other incremental learning algorithms on the Permuted-MNIST data with only three epochs (as compared with ten epochs for all the other algorithms). The naive rehearsal approach outperforms all the other approaches, but it  comes with the additional memory overhead.

\subsection{Transferability Study} 
Identification of the best configuration for each group of classes (a dataset) 
is based on the assumption that all the data is available at the beginning of 
the training procedure. However,  for many online learning
scenarios, both continual and single task learning, this 
configuration might not be known \emph{a priori}. 
This raises the question of transfer metalearning: how to
effectively optimize a learning 
algorithm to learn unknown tasks and data.
To address this problem, we empirically
study the effect of transferring optimal configurations
tasks for the continual learning scenario,
and across datasets for the single task learning case. 

In Table~\ref{tab:Transfer-datasets} we show the accuracy resulting from transferring
optimal conditions across tasks and across datasets. From these results we can
extract several conclusions: first, transfer from single task online to task-incremental
continual learning is excellent for the MSE rule, but not for the inelastic rule. We
attribute this effect to the strong role that synaptic weight distribution has on the inelastic rule. Second, for the class-incremental case, as already seen in Table 4, the accuracies obtained
when transferring hyperparameters are significantly lower than those achieved when the conditions are specifically optimized for  class-incremental learning. This indicates that, at least in memory-constrained situations, learning rules have to be highly optimized to class-incremental problems, and that we have to rely on surrogate data to carry out transfer metalearning.

\begin{table*}[t]
\caption{Accuracy due to transfer metalearning from single task online learning to task-incremental and and class-incremental continual learning experiments.} 
\begin{center}
\begin{adjustbox}{max width=\textwidth}
\begin{tabular}{|c|c|c H |c|c|c|c|c|c|}
 \hline
      & \multicolumn{4}{c|}{Single Task Learning (0.5 epoch)} & &\multicolumn{2}{c|}{Task-Incremental Learning} &\multicolumn{2}{c|}{Class-Incremental Learning}\\
    \hline
      Config ($\downarrow$), Dataset ($\rightarrow$)  & MNIST  & F-MNIST  & E-MNIST & CIFAR10 & & SplitMNIST & SplitCIFAR-10 & SplitMNIST & SplitCIFAR-10\\
    \hline 
    \hline
      MNIST (w/INEL) & $96.39$  & $85.48$ & -- & $72.91$ & & $65.25$ & $77.63$ & $21.56$ & $21.19$\\
    \hline
      MNIST (w/MSE) & $96.16$   & $80.75$  & -- & $69.87$ & & $99.60$ & $92.50$ & $21.84$ & $20.33$\\
    \hline
      F-MNIST & $94.52$  & $85.13$  & -- & $71.29$ & & $99.25$ & $94.37$ & $21.39$ & $21.19$\\
    \hline
      CIFAR-10 (w/INEL)& $93.99$   & $84.13$  & -- & $77.95$ & & $64.99$ & $81.20$ & $20.86$ & $22.33$\\
    \hline
      CIFAR-10 (w/MSE) & $94.45$  & $84.21$  & -- & $77.37$ & & $99.03$ & $94.65$ & $21.75$ & $22.85$\\
    \hline

\end{tabular}
\end{adjustbox}
\end{center}
\label{tab:Transfer-datasets}
\end{table*}

\noindent{\bf Distance metrics as a measure of transferability:}
To rationalize the results shown in Table \ref{tab:Transfer-datasets}, we have explored the correlation
between performance drop across datasets and their
effective dimensionality and the distance between datasets.
We consider as a metric of
dimensionality the sum of squares of the eigenvalues ($\lambda$) of the covariance matrix of the complete dataset, so that
\begin{equation}
    d = (\sum_i\lambda_i)^2/\sum_i \lambda_i^2 .
    \label{eq:eigen-dim}
\end{equation}
This metric
has been used in the past to characterize sparse representations \cite{Litwin_2017}.

\begin{table}[ht!]
\caption{PCA metric for each of the datasets 
(Eq. \ref{eq:eigen-dim})}
\begin{center}
\begin{adjustbox}{max width=0.48\textwidth}
\begin{tabular}{|c|c|c|c| c|}
    \hline
      Data  & MNIST      & F-MNIST    & E-MNIST    & CIFAR-10 \\
    \hline 
      Eigen-Dim   & $30.69$ & $7.91$ & $27.09$ & $132.60$ \\
    \hline
\end{tabular}
\end{adjustbox}
\end{center}
\label{tab:pcadim}
\end{table}

We calculate the distance between two datasets using the minimum cosine distance between any two categories.
The results are show in  
Table~\ref{tab:distances},  where we also show the maximum separation
for comparison. Since cosine distances can be applied only to vectors with the same dimensionality, we cannot include the CIFAR-10 dataset in this analysis.

\begin{figure}[t!]
\centering
  \includegraphics[width=0.95\linewidth]{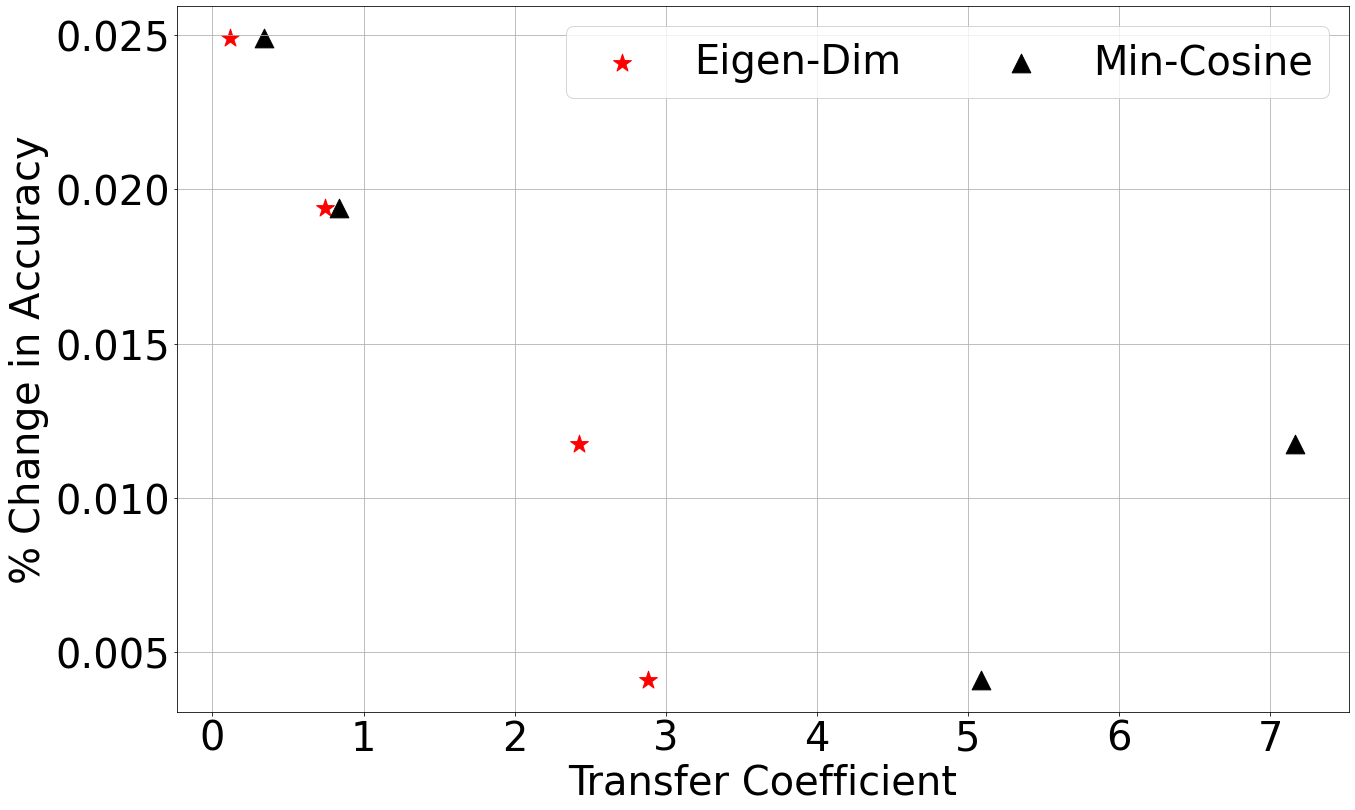}
\caption{Drop in classification accuracy  during  transfer  metalearning as a function of a transfer coefficient obtained for PCA metric and minimum cosine distance metric.}%
\label{fig:trnsfr}
\end{figure}

Figure \ref{fig:trnsfr} shows the 
drop in classification accuracy during transfer metalearning
as a function of a \emph{transfer coefficient}
obtained from two different metrics: the relative
difference in the eigenvalue
dimension given by Eq. \ref{eq:eigen-dim} and the minimum
cosine distance between the two datasets. Both values
are normalized as
\begin{equation}
\label{eq:transfer}
D = {|{M_1-M_2}|}/{M_1},
\end{equation}
where $M_1$ is the metric of the dataset
selected to run the experiment and $M_2$ is the metric
of the dataset whose optimal configuration is used. Note that this
transfer coefficient is not symmetric because of the different
normalization in Eq. \ref{eq:transfer}, as should be expected since  transfer
metalearning is directional. The results clearly show that as the distance between the datasets increases, transfer learning the configurations across them leads to a decrease in accuracy.

\begin{table}
\centering
\caption{Minimum and maximum cosine distances
    between centroids of the categories of different datasets}
\begin{adjustbox}{max width=0.48\textwidth}
\begin{tabular}{|c|c|c|c| H|}

    \hline
      Data  & MNIST      & F-MNIST    & E-MNIST  \\
    \hline 
      MNIST & $(0.073,0.55)$ & $(0.17,0.54)$ & $(0.044,0.59)$ \\
    \hline
      F-MNIST &          & $(0.012,0.56)$ & $(0.13,0.57)$ \\
    \hline
      E-MNIST &        &          &   $(0.098,0.52)$  \\
    \hline
     
\end{tabular}
\end{adjustbox}
\label{tab:distances}
\end{table}

\section{Summary and Conclusions}

In this work we have applied bio-inspired design principles to the problem of memory-free online continual learning. Our architecture relies on local learning rules to carry out single task online and continual learning tasks. By formalizing the different components we have been able to employ a Bayesian optimization-based search to find optimal task-specific configurations over a mixed (categorical, continuous) space spanning learning algorithms, feature extraction layers, and their hyperparameters. We have used this methodology both to identify optimal configurations for each case and to explore transfer metalearning across modalities and datasets.

 By systematically 
transferring the configurations metalearned in the single task online learning scenario to task-incremental continual learning, we obtained accuracies exceeding 99\% and 95\% for 
Split-MNIST and Split-CIFAR-10, both
exceeding the state-of-the-art of 
memory-based and memory-free algorithms. 
In class-incremental learning,
our architecture clearly outperforms memory-free algoritms in the Split-MNIST and Split-CIFAR-10 tests. It achieves an accuracy in Split-CIFAR-10 of 55.74 that also exceeds that of memory-based algorithms. However, in contrast to task-incremental learning case, we were not able to achieve this performance with configurations transferred from the online single task case, but they had to be specifically optimized for the class-incremental tests.

An important fraction of the highest performing configurations identified in this work relied on a novel inelastic rule implementing a simple form of memory consolidation for synaptic weights that deviate from the pool of presynaptic weights of each neuron (Eq. 12). This rule leads to the long-term stabilization of weights that are relevant for a specific class, mitigating catastrophic forgetting. This mechanism can be easily combined with any synaptic plasticity mechanism to stabilize networks learning from non-stationary data. Configurations using this rule achieved accuracies that are three times higher than the best results obtained using memory-free algorithms in the class-incremental tasks considered in this work. This highlights the potential of moving beyond conventional local learning rules to fully realize the possibilities afforded by recurrent neural networks to control learning and plasticity in feedforward neural networks.

\bibliography{main}

\begin{thebibliography}{40}
\providecommand{\natexlab}[1]{#1}
\providecommand{\url}[1]{\texttt{#1}}
\expandafter\ifx\csname urlstyle\endcsname\relax
  \providecommand{\doi}[1]{doi: #1}\else
  \providecommand{\doi}{doi: \begingroup \urlstyle{rm}\Url}\fi

\bibitem[Aljundi et~al.(2019{\natexlab{a}})Aljundi, Caccia, Belilovsky, Caccia,
  Lin, Charlin, and Tuytelaars]{aljundi2019online}
Aljundi, R., Caccia, L., Belilovsky, E., Caccia, M., Lin, M., Charlin, L., and
  Tuytelaars, T.
\newblock Online continual learning with maximally interfered retrieval.
\newblock \emph{arXiv preprint arXiv:1908.04742}, 2019{\natexlab{a}}.

\bibitem[Aljundi et~al.(2019{\natexlab{b}})Aljundi, Lin, Goujaud, and
  Bengio]{aljundi2019gradient}
Aljundi, R., Lin, M., Goujaud, B., and Bengio, Y.
\newblock Gradient based sample selection for online continual learning.
\newblock \emph{arXiv preprint arXiv:1903.08671}, 2019{\natexlab{b}}.

\bibitem[{Balaprakash} et~al.(2018){Balaprakash}, {Salim}, {Uram},
  {Vishwanath}, and {Wild}]{Balaprakash_DH_2018}
{Balaprakash}, P., {Salim}, M., {Uram}, T., {Vishwanath}, V., and {Wild}, S.
\newblock Deephyper: Asynchronous hyperparameter search for deep neural
  networks.
\newblock In \emph{2018 IEEE 25th International Conference on High Performance
  Computing (HiPC)}, pp.\  42--51, Dec 2018.
\newblock \doi{10.1109/HiPC.2018.00014}.

\bibitem[Beaulieu et~al.(2020)Beaulieu, Frati, Miconi, Lehman, Stanley, Clune,
  and Cheney]{beaulieu2020learning}
Beaulieu, S., Frati, L., Miconi, T., Lehman, J., Stanley, K.~O., Clune, J., and
  Cheney, N.
\newblock Learning to continually learn.
\newblock \emph{arXiv preprint arXiv:2002.09571}, 2020.

\bibitem[Buzzega et~al.(2020)Buzzega, Boschini, Porrello, Abati, and
  Calderara]{buzzega2020dark}
Buzzega, P., Boschini, M., Porrello, A., Abati, D., and Calderara, S.
\newblock Dark experience for general continual learning: a strong, simple
  baseline.
\newblock \emph{arXiv preprint arXiv:2004.07211}, 2020.

\bibitem[Chaudhry et~al.(2018)Chaudhry, Ranzato, Rohrbach, and
  Elhoseiny]{chaudhry2018efficient}
Chaudhry, A., Ranzato, M., Rohrbach, M., and Elhoseiny, M.
\newblock Efficient lifelong learning with a-gem.
\newblock \emph{arXiv preprint arXiv:1812.00420}, 2018.

\bibitem[Chen et~al.(2020{\natexlab{a}})Chen, Cheng, Juan, Wei, and
  Sun]{InstaParam}
Chen, H.-J., Cheng, A.-C., Juan, D.-C., Wei, W., and Sun, M.
\newblock Mitigating forgetting in online continual learning via instance-aware
  parameterization.
\newblock In \emph{Advances in Neural Information Processing Systems},
  volume~33, 2020{\natexlab{a}}.

\bibitem[Chen et~al.(2020{\natexlab{b}})Chen, Cheng, Juan, Wei, and
  Sun]{chen2020mitigating}
Chen, H.-J., Cheng, A.-C., Juan, D.-C., Wei, W., and Sun, M.
\newblock Mitigating forgetting in online continual learning via instance-aware
  parameterization.
\newblock \emph{Advances in Neural Information Processing Systems}, 33,
  2020{\natexlab{b}}.

\bibitem[Cohen et~al.(2017)Cohen, Afshar, Tapson, and
  Van~Schaik]{cohen2017emnist}
Cohen, G., Afshar, S., Tapson, J., and Van~Schaik, A.
\newblock Emnist: Extending mnist to handwritten letters.
\newblock In \emph{2017 International Joint Conference on Neural Networks
  (IJCNN)}, pp.\  2921--2926. IEEE, 2017.

\bibitem[Deng et~al.(2009)Deng, Dong, Socher, Li, Li, and
  Fei-Fei]{deng2009imagenet}
Deng, J., Dong, W., Socher, R., Li, L.-J., Li, K., and Fei-Fei, L.
\newblock Imagenet: A large-scale hierarchical image database.
\newblock In \emph{2009 IEEE conference on computer vision and pattern
  recognition}, pp.\  248--255. Ieee, 2009.

\bibitem[Farquhar \& Gal(2018)Farquhar and Gal]{farquhar2018towards}
Farquhar, S. and Gal, Y.
\newblock Towards robust evaluations of continual learning.
\newblock \emph{arXiv preprint arXiv:1805.09733}, 2018.

\bibitem[Hige et~al.(2015)Hige, Aso, Modi, Rubin, and
  Turner]{Hige_nonhebbian_2015}
Hige, T., Aso, Y., Modi, M.~N., Rubin, G.~M., and Turner, G.~C.
\newblock Heterosynaptic plasticity underlies aversive olfactory learning in
  <em>drosophila</em>.
\newblock \emph{Neuron}, 88\penalty0 (5):\penalty0 985--998, 2020/07/05 2015.
\newblock \doi{10.1016/j.neuron.2015.11.003}.
\newblock URL \url{https://doi.org/10.1016/j.neuron.2015.11.003}.

\bibitem[Hsu et~al.(2018)Hsu, Liu, Ramasamy, and Kira]{hsu2018re}
Hsu, Y.-C., Liu, Y.-C., Ramasamy, A., and Kira, Z.
\newblock Re-evaluating continual learning scenarios: A categorization and case
  for strong baselines.
\newblock \emph{arXiv preprint arXiv:1810.12488}, 2018.

\bibitem[Javed \& White(2019)Javed and White]{javed2019meta}
Javed, K. and White, M.
\newblock Meta-learning representations for continual learning.
\newblock In \emph{Advances in Neural Information Processing Systems}, pp.\
  1818--1828, 2019.

\bibitem[Kirkpatrick et~al.(2017)Kirkpatrick, Pascanu, Rabinowitz, Veness,
  Desjardins, Rusu, Milan, Quan, Ramalho, Grabska-Barwinska,
  et~al.]{kirkpatrick2017overcoming}
Kirkpatrick, J., Pascanu, R., Rabinowitz, N., Veness, J., Desjardins, G., Rusu,
  A.~A., Milan, K., Quan, J., Ramalho, T., Grabska-Barwinska, A., et~al.
\newblock Overcoming catastrophic forgetting in neural networks.
\newblock \emph{Proceedings of the national academy of sciences}, 114\penalty0
  (13):\penalty0 3521--3526, 2017.

\bibitem[Lee et~al.(2020)Lee, Ha, Zhang, and Kim]{Lee2020A}
Lee, S., Ha, J., Zhang, D., and Kim, G.
\newblock A neural dirichlet process mixture model for task-free continual
  learning.
\newblock In \emph{International Conference on Learning Representations}, 2020.
\newblock URL \url{https://openreview.net/forum?id=SJxSOJStPr}.

\bibitem[Li \& Hoiem(2017)Li and Hoiem]{li2017learning}
Li, Z. and Hoiem, D.
\newblock Learning without forgetting.
\newblock \emph{IEEE transactions on pattern analysis and machine
  intelligence}, 40\penalty0 (12):\penalty0 2935--2947, 2017.

\bibitem[Lindauer et~al.(2019)Lindauer, Eggensperger, Feurer, Biedenkapp,
  Marben, Müller, and Hutter]{CondConstraint2019}
Lindauer, M., Eggensperger, K., Feurer, M., Biedenkapp, A., Marben, J.,
  Müller, P., and Hutter, F.
\newblock Boah: A tool suite for multi-fidelity bayesian optimization $\&$
  analysis of hyperparameters.
\newblock \emph{arXiv:1908.06756 {[cs.LG]}}, 2019.

\bibitem[Litwin-Kumar et~al.(2017)Litwin-Kumar, Harris, Axel, Sompolinsky, and
  Abbott]{Litwin_2017}
Litwin-Kumar, A., Harris, K.~D., Axel, R., Sompolinsky, H., and Abbott, L.~F.
\newblock Optimal degrees of synaptic connectivity.
\newblock \emph{Neuron}, 93\penalty0 (5):\penalty0 1153--1164.e7, 2020/06/10
  2017.
\newblock \doi{10.1016/j.neuron.2017.01.030}.
\newblock URL \url{https://doi.org/10.1016/j.neuron.2017.01.030}.

\bibitem[Lopez-Paz \& Ranzato(2017)Lopez-Paz and Ranzato]{lopez2017gradient}
Lopez-Paz, D. and Ranzato, M.
\newblock Gradient episodic memory for continual learning.
\newblock In \emph{Advances in Neural Information Processing Systems}, pp.\
  6467--6476, 2017.

\bibitem[Mai et~al.(2021)Mai, Li, Jeong, Quispe, Kim, and
  Sanner]{mai2021online}
Mai, Z., Li, R., Jeong, J., Quispe, D., Kim, H., and Sanner, S.
\newblock Online continual learning in image classification: An empirical
  survey.
\newblock \emph{arXiv preprint arXiv:2101.10423}, 2021.

\bibitem[Miconi et~al.(2018)Miconi, Clune, and Stanley]{Miconi2018}
Miconi, T., Clune, J., and Stanley, K.~O.
\newblock Differentiable plasticity: training plastic neural networks with
  backpropagation.
\newblock \emph{CoRR}, abs/1804.02464, 2018.
\newblock URL \url{http://arxiv.org/abs/1804.02464}.

\bibitem[Nguyen et~al.(2017)Nguyen, Li, Bui, and Turner]{nguyen2017variational}
Nguyen, C.~V., Li, Y., Bui, T.~D., and Turner, R.~E.
\newblock Variational continual learning.
\newblock \emph{arXiv preprint arXiv:1710.10628}, 2017.

\bibitem[Oja(1982)]{Oja1982}
Oja, E.
\newblock Simplified neuron model as a principal component analyzer.
\newblock \emph{Journal of Mathematical Biology}, 15\penalty0 (3):\penalty0
  267--273, 1982.
\newblock \doi{10.1007/BF00275687}.
\newblock URL \url{https://doi.org/10.1007/BF00275687}.

\bibitem[Parisi et~al.(2019)Parisi, Kemker, Part, Kanan, and
  Wermter]{parisi2019continual}
Parisi, G.~I., Kemker, R., Part, J.~L., Kanan, C., and Wermter, S.
\newblock Continual lifelong learning with neural networks: A review.
\newblock \emph{Neural Networks}, 2019.

\bibitem[Pogodin \& Latham(2020)Pogodin and Latham]{pogodin2020kernelized}
Pogodin, R. and Latham, P.~E.
\newblock Kernelized information bottleneck leads to biologically plausible
  3-factor hebbian learning in deep networks.
\newblock \emph{arXiv preprint arXiv:2006.07123, NeurIPS 2020}, 2020.

\bibitem[Rajasegaran et~al.(2019)Rajasegaran, Hayat, Khan, Khan, and
  Shao]{NEURIPS2019_83da7c53}
Rajasegaran, J., Hayat, M., Khan, S.~H., Khan, F.~S., and Shao, L.
\newblock Random path selection for continual learning.
\newblock In Wallach, H., Larochelle, H., Beygelzimer, A., d\textquotesingle
  Alch\'{e}-Buc, F., Fox, E., and Garnett, R. (eds.), \emph{Advances in Neural
  Information Processing Systems}, volume~32, pp.\  12669--12679. Curran
  Associates, Inc., 2019.

\bibitem[Rajasegaran et~al.(2020)Rajasegaran, Khan, Hayat, Khan, and
  Shah]{rajasegaran2020itaml}
Rajasegaran, J., Khan, S., Hayat, M., Khan, F.~S., and Shah, M.
\newblock itaml: An incremental task-agnostic meta-learning approach.
\newblock \emph{arXiv preprint arXiv:2003.11652}, 2020.

\bibitem[Rebuffi et~al.(2017)Rebuffi, Kolesnikov, Sperl, and
  Lampert]{rebuffi2017icarl}
Rebuffi, S.-A., Kolesnikov, A., Sperl, G., and Lampert, C.~H.
\newblock icarl: Incremental classifier and representation learning.
\newblock In \emph{Proceedings of the IEEE conference on Computer Vision and
  Pattern Recognition}, pp.\  2001--2010, 2017.

\bibitem[Rusu et~al.(2016)Rusu, Rabinowitz, Desjardins, Soyer, Kirkpatrick,
  Kavukcuoglu, Pascanu, and Hadsell]{rusu2016progressive}
Rusu, A.~A., Rabinowitz, N.~C., Desjardins, G., Soyer, H., Kirkpatrick, J.,
  Kavukcuoglu, K., Pascanu, R., and Hadsell, R.
\newblock Progressive neural networks.
\newblock \emph{arXiv preprint arXiv:1606.04671}, 2016.

\bibitem[Schwarz et~al.(2018)Schwarz, Czarnecki, Luketina, Grabska-Barwinska,
  Teh, Pascanu, and Hadsell]{schwarz2018progress}
Schwarz, J., Czarnecki, W., Luketina, J., Grabska-Barwinska, A., Teh, Y.~W.,
  Pascanu, R., and Hadsell, R.
\newblock Progress \& compress: A scalable framework for continual learning.
\newblock In \emph{International Conference on Machine Learning}, pp.\
  4528--4537, 2018.

\bibitem[Shin et~al.(2017)Shin, Lee, Kim, and Kim]{shin2017continual}
Shin, H., Lee, J.~K., Kim, J., and Kim, J.
\newblock Continual learning with deep generative replay.
\newblock In \emph{Advances in Neural Information Processing Systems}, pp.\
  2990--2999, 2017.

\bibitem[van~de Ven \& Tolias(2018)van~de Ven and Tolias]{van2018generative}
van~de Ven, G.~M. and Tolias, A.~S.
\newblock Generative replay with feedback connections as a general strategy for
  continual learning.
\newblock \emph{arXiv preprint arXiv:1809.10635}, 2018.

\bibitem[Xiao et~al.(2017{\natexlab{a}})Xiao, Rasul, and
  Vollgraf]{xiao2017_online}
Xiao, H., Rasul, K., and Vollgraf, R.
\newblock Fashion-mnist: a novel image dataset for benchmarking machine
  learning algorithms.
\newblock \emph{arXiv preprint arXiv:1708.07747}, 2017{\natexlab{a}}.

\bibitem[Xiao et~al.(2017{\natexlab{b}})Xiao, Rasul, and
  Vollgraf]{xiao2017fashion}
Xiao, H., Rasul, K., and Vollgraf, R.
\newblock Fashion-mnist: a novel image dataset for benchmarking machine
  learning algorithms.
\newblock \emph{arXiv preprint arXiv:1708.07747}, 2017{\natexlab{b}}.

\bibitem[Yanguas-Gil(2019)]{ayg_memristors}
Yanguas-Gil, A.
\newblock Memristor design rules for dynamic learning and edge processing
  applications.
\newblock \emph{APL Materials}, 7\penalty0 (9):\penalty0 091102, 2019.
\newblock \doi{10.1063/1.5109910}.
\newblock URL \url{https://doi.org/10.1063/1.5109910}.

\bibitem[{Yanguas-Gil} et~al.(2019){Yanguas-Gil}, {Mane}, {Elam}, {Wang},
  {Severa}, {Daram}, and {Kudithipudi}]{YanguasGil_MSE_2019}
{Yanguas-Gil}, A., {Mane}, A., {Elam}, J.~W., {Wang}, F., {Severa}, W.,
  {Daram}, A.~R., and {Kudithipudi}, D.
\newblock The insect brain as a model system for low power electronics and edge
  processing applications.
\newblock In \emph{2019 IEEE Space Computing Conference (SCC)}, pp.\  60--66,
  2019.

\bibitem[Zagoruyko \& Komodakis(2016)Zagoruyko and
  Komodakis]{zagoruyko2016wide}
Zagoruyko, S. and Komodakis, N.
\newblock Wide residual networks.
\newblock \emph{arXiv preprint arXiv:1605.07146}, 2016.

\bibitem[Zenke et~al.(2017)Zenke, Poole, and Ganguli]{zenke2017continual}
Zenke, F., Poole, B., and Ganguli, S.
\newblock Continual learning through synaptic intelligence.
\newblock In \emph{Proceedings of the 34th International Conference on Machine
  Learning-Volume 70}, pp.\  3987--3995. JMLR. org, 2017.

\bibitem[Zhang et~al.(2020)Zhang, Zhang, Ghosh, Li, Tasci, Heck, Zhang, and
  Kuo]{zhang2020class}
Zhang, J., Zhang, J., Ghosh, S., Li, D., Tasci, S., Heck, L., Zhang, H., and
  Kuo, C.-C.~J.
\newblock Class-incremental learning via deep model consolidation.
\newblock In \emph{Proceedings of the IEEE/CVF Winter Conference on
  Applications of Computer Vision}, pp.\  1131--1140, 2020.

\end{thebibliography}
\bibliographystyle{icml2021}

\newpage
\appendix




\section{ Mixed-Integer Black-box Optimization procedure}

All the single-task and class-incremental task optimization has been carried out using 
DeepHyper~\cite{Balaprakash_DH_2018}, which is an open source optimization library. All the optimization experiments
utilized $128$ nodes and a wall time of $90$ minutes. We also note that since all the four learning 
rules adopted in this work (GEN, OJA, MSE, INEL) have different number of tunable parameters, we 
utilize conditional constraints~\cite{CondConstraint2019} to  keep the search space consistent across 
the learning rules and to improve the optimization efficiency. The conditional constraints only 
keep the relevant parameters of a learning rule active while setting the rest to NaN during the 
optimization.

\section{Impact of configuration on catastrophic forgetting}

To further understand why the configurations learning in the single-task scenario transferred well to the 
Task-Incremental scenario but not to the Class-incremental case, we studied
the evolution with time of the weight matrix ($W_l$) in the
neuromodulated learning layer,
which has dimensions  of $W_l$ of $x_e \times x_o$.

More specifically, we tracked the evolution
of the mean weight presynaptic to each
output neuron: by averaging $W_l$ across
 $x_e$, we produce a $x_o$
dimensional array for every data sample streamed to the model during the learning. 
We use this to track learning
for each class. 

\begin{figure}[h!]
\centering
  \includegraphics[width=\linewidth]{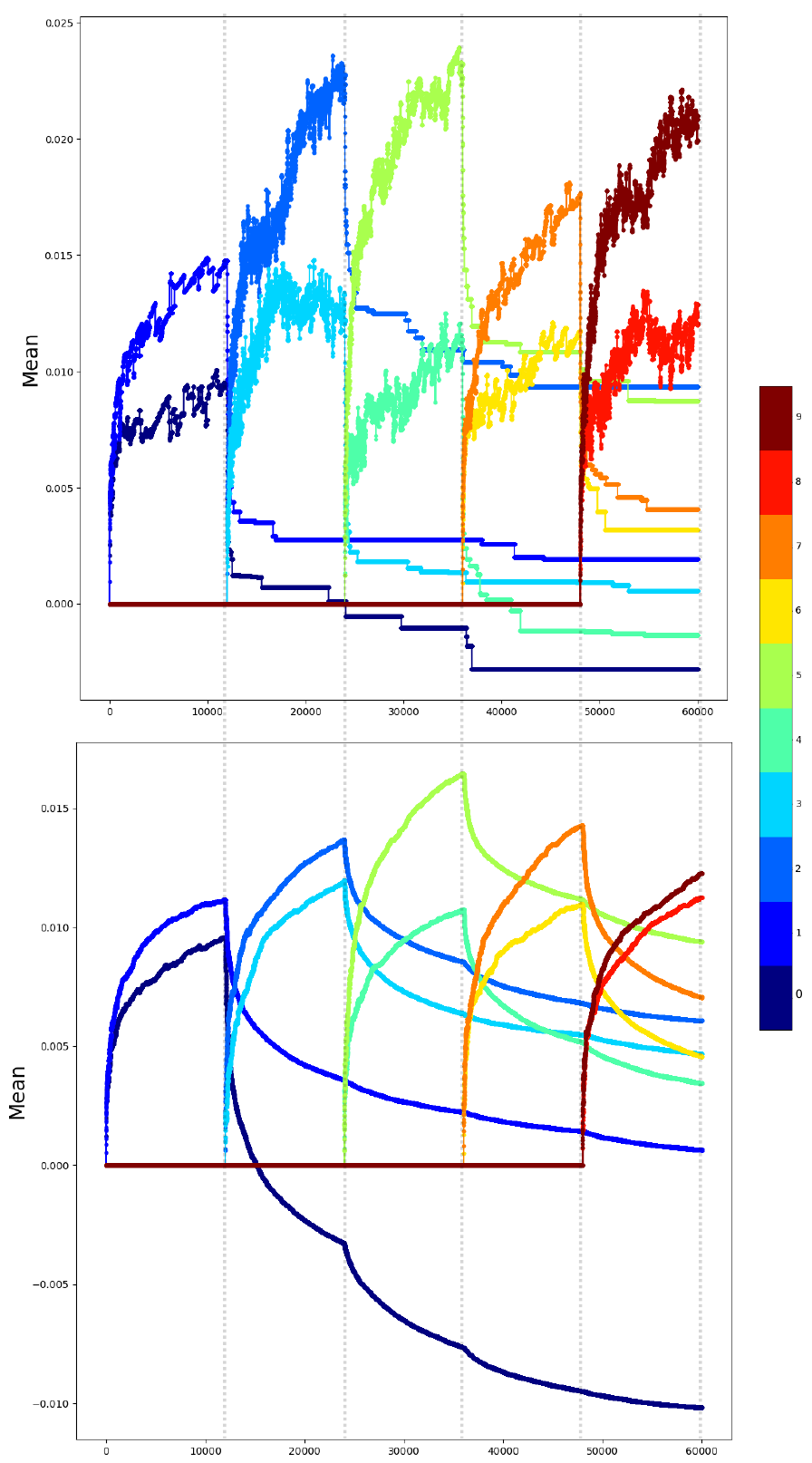}
\caption{Comparison of the weights in the neuromodulated learning layer for split MNIST learned in the class-incremental learning scenario using the configuration from single task learning (transfer w/ INEL) and that optimized for the class incremental learning (Opt w/ Split MNIST).} 
\label{fig:SMNIST_Cinc}
\end{figure}

The evolution of this 
mean weight is shown in Figure~\ref{fig:SMNIST_Cinc}
for the Split-MNIST data in the class-incremental learning scenario.
The top plot corresponds to the
INEL configuration transferred from 
the single task learning, while the 
bottom plot represents the
evolution of the mean synaptic weight for the configuration learned by optimizing over the class-incremental learning itself. Each
task is composed of 12k samples.

After a specific task is concluded,
the mean synaptic weights for the classes
learned during that task suffer a
significant drop, which is consistent
with the presence of catastrophic forgetting
and lower accuracies. The magnitude
of this drop, however, is significantly
more gradual in the configuration
optimized for the continual learning
case. This correlates with
the good class incremental learning accuracy observed with this configuration. This
more gradual decay is a consequence of
the inelasticity of the learning rate, which remains zero for weights that differ
significantly from the mean in the
INEL rule.

To evaluate the transferability of the
configuration obtained from
optimizing the class-incremental learning
across datasets, we applied this
very same configuration, learned with
on Split-MNIST, to the Split-CIFAR-10
and the Split-CIFAR-100 datasets.
The resulting accuracies,
$45.95$ for Split-CIFAR-10 and $25.56$
for Split-CIFAR-100,  outperform most
of the memory-free and the 
memory-based methods we compared against (Table \ref{tab:TaskINC-offline-expand}).

\begin{table*}[t]
\caption{Optimal configurations obtained for Split-MNIST, and Split-CIFAR-10 datasets using the mixed-integer black-box optimization in the class-incremental learning scenario, where each configuration was evaluated after $1.0$ epochs of training.}
\begin{center}
\begin{adjustbox}{max width=\textwidth}
\begin{tabular}{|c|c|c|c|c|c|c|c|c|c|c|c|c|c|}
    \hline
    Dataset & Acc. &  Srule & $N_H^{F}$ & $\beta_1^{F}$& $\beta_2^{F}$ & $\beta_3^{F}$ & $\gamma^{F}$ &  $\beta_1^{M}$ & $\beta_2^{M}$ & $\beta_3^{M}$ & $\gamma^{M}$ & $\alpha^{M}$  \\
    \hline
    \hline
    MNIST & $78.77$ & INEL & $11000$ & $0.143$ & $0.501$ & $1.861$ & -- &  -- & $0.954$ & -- & $0.094$ & --\\
    \hline
    CIFAR-10 & $55.75$ & INEL & $11000$ & $0.011$ & $0.103$ & $0.137$ & --  & -- & $0.920$ & -- & $0.293$ & --  \\
    \hline
\end{tabular}
\end{adjustbox}
\end{center}
\label{tab:CLinc-acc}
\end{table*}

\begin{table*}[ht!]
\caption{Classification accuracy for the task-incremental and class-incremental learning experiments on Split-MNIST, Split CIFAR-10, and Split CIFAR-100 datasets. We employ three configurations of our approach, the configurations corresponding to that of INEL rule, MSE rule learned in the single task online learning. The third configuration is explicitly learned through optimization in the class-incremental learning scenario. The accuracy metrics are reported as mean and standard deviation over $5$ repetitions. (Note: This table is an extension for Table~\ref{tab:TaskINC-offline} presented in the paper, where only accuracy for a single run with Split-MNIST and Split-CIFAR10 have been reported.) } 
\begin{center}
\begin{adjustbox}{max width=\linewidth}
\begin{tabular}{|c|c|c|c|c|c|c|c|}
 \hline
      &  &\multicolumn{3}{c|}{Task-Incremental Learning} &\multicolumn{3}{c|}{Class-Incremental Learning}\\
    \hline
          & Method & Split-MNIST & Split-CIFAR-10 & Split-CIFAR-100 & Split-MNIST & Split-CIFAR-10 & Split-CIFAR-100\\ 
    \hline
    \hline
    \multirow{2}{*}{Baseline}& iid-offline  & $99.46 \pm 0.08$ & $95.51 \pm 0.22$  & $80.89 \pm 0.75$ & $95.82 \pm 0.33$ &  $80.54 \pm 0.63$ & $48.092 \pm 0.90$\\
    & Fine-Tune  &  $97.29 \pm 0.96$ & $62.56 \pm 5.76$  & $47.54 \pm 1.61$ & $19.68 \pm 0.02$ & $19.19 \pm 0.06 $ &  $8.32 \pm 0.23$\\
    \hline
    \hline
    \multirow{3}{*}{\shortstack[l]{Continual Learning \\ Memory-free }}& Online EWC  &$98.46 \pm 0.47$ & $59.98 \pm 2.27$  & $20.24 \pm 1.23$ & $19.92 \pm 0.35$ & $16.18 \pm 1.37$ & $4.41 \pm 0.37$\\
     &SI&  $97.95 \pm 0.70$ & $65.71 \pm 1.79$  & $33.00 \pm 2.59$ & $19.76 \pm 0.01$ & $17.27 \pm 0.87$ & $5.87 \pm 0.21$ \\
     &LwF&  $99.26 \pm 0.09$& $63.47 \pm 1.52$  & $19.45 \pm 1.19$  & $20.54 \pm 0.64$ & $18.53 \pm 0.1$2 & $6.93 \pm 0.32$\\
     &InstAParam  & -- & 83.8 &  55.5 & -- & -- &  --\\
     \cline{1-8}
     \multirow{8}{*}{\shortstack[l]{Continual Learning \\ Memory-based \\ (Buffer=0.5k) }}&A-GEM  &  $99.16 \pm 0.08$ & $72.02 \pm 1.29$  & $38.39 \pm 1.98$ & $48.57 \pm 5.26$ & $18.21 \pm 0.16$ & $6.18 \pm 0.20$\\
     &iCaRL  &  $98.40 \pm 0.09$  & $82.01 \pm 0.76$  & $50.56 \pm 0.23$ & $72.55 \pm 0.45$ & $35.88 \pm 1.43$ & $15.76 \pm 0.15$\\
     &GSS  &  $97.80 \pm 0.80$  & $86.38 \pm 1.21$  & $56.86 \pm 1.68$ & $54.14 \pm 4.68$ & $49.22 \pm 1.71$  & $11.33 \pm 0.40$\\
     &RPSNet & -- & 67.0   & 40.1 & -- & -- &  --\\

     & ER-MIR  & -- & --  & -- & $86.60 \pm 1.60$ & $37.80 \pm 1.80$ & $9.20 \pm 0.40$\\
     & CN-DPM & -- & --  & -- & ${\bf 93.81}\pm 0.07$ & $47.05 \pm 0.62$ & $16.13 \pm 0.14$ \\
     &DER++  &  $99.29 \pm 0.02$ & $85.95 \pm 1.62$  & $58.28 \pm 1.50$ & $92.21 \pm 0.54$ & $52.01 \pm 3.06$ & $15.04 \pm 1.044$\\
     \cline{1-8}
     & Ours (transfer w/ INEL) &  $65.13 \pm 0.66$  & $81.05 \pm 0.15$  & $23.04 \pm 0.26$ & $21.39 \pm 0.23$ & $21.93 \pm 0.38$ &  $11.95 \pm 0.20$ \\
     & Ours (transfer w/ MSE ) &  ${\bf 99.56} \pm 0.04$ & ${\bf 94.43} \pm 0.15$  & ${\bf 83.17} \pm 0.15$ & $21.22 \pm 0.79$ & $22.37 \pm 0.28$ & $12.25 \pm 0.31$\\
     & Ours (Opt w/ Split MNIST) &  --  & --   & -- & $77.25 \pm 1.02$ & $45.95 \pm 0.90$ & ${\bf 25.56} \pm 0.69$ \\
     & Ours (Opt  w/ Split CIFAR10) &  --  & --   & -- & $60.82 \pm 2.00$ & ${\bf 52.55} \pm 2.05$ & $18.87 \pm 0.33$ \\

    \hline
\end{tabular}
\end{adjustbox}
\end{center}
\label{tab:TaskINC-offline-expand}
\end{table*}

\section{Additional Experiments}
All the memory-based continual experiments in Table~\ref{tab:TaskINC-offline}  use a buffer size 
of 0.5K samples, batch size=10 and epochs=1. To study the performance of our approach in case of 
experiments with higher number of tasks and classes per task, we consider the CIFAR-100 dataset 
following other works~\cite{mai2021online,Lee2020A,chen2020mitigating}. We split the CIFAR-100 dataset such that
each task consists of 10 consecutive classes, thus leading to a total of 10 tasks.

Consistent with the experiments in the main text, the Split MNIST use a multi layered perceptron model 
with two layers and $400$ nodes per layer, while Split CIFAR-10 and Split CIFAR-100 experiments 
use as ResNet-18 model.
The experiments for Split MNIST, Split CIFAR-10 and Split CIFAR-100 datasets using the models 
iid-offline, Fine-Tune, Online EWC, SI, LwF, A-GEM, iCaRL, GSS, DER++ has been run by adopting 
the code in \url{https://github.com/aimagelab/mammoth}. Experiments with ER-MIR have been run 
by adopting the code in \url{https://github.com/optimass/Maximally_Interfered_Retrieval}. The 
experiments with CN-DPM adopted the implementation in \url{https://github.com/soochan-lee/CN-DPM}.
In all the above models, we ensured that the hyperparameters used are conistent with the 
online continual learning scenario mentioned above. For InstAParam,  RPSNet, we report the 
metrics from the paper~\cite{chen2020mitigating}, since the experiments were run with the same 
hyperparameters as those used for other reported approaches in this work.

\section{Ablation Studies}
In the proposed architecture consisting of the feature extraction layer and the neuromodulated 
learning layer, we study the importance of each component by performing an ablation study. We 
fix the dataset to MNIST and run it in the single-task learning scenario with the MNIST-INEL 
configuration shown in Table~\ref{tab:Ninc-acc} for these experiments. In the first experiment,
the neuromodulated learning layer is replaced by a Multi-Layered Perceptron model with two layers,
each with 400 nodes, while keeeping the feature extraction layer intact. In this scenario, we found
a drop in test accuracy to 37$\%$ (with 10 epochs) from the $96.4\%$ obtained with the proposed architecture. Next, we
consider both the feature extraction and the neuromodulated learning layers, but turn off the sparsity
imposed by the dynamic thresholding (Equ.~\ref{eq:dyn_thresh}). This lead to an accuracy drop to $23.24\%$,
which signifies the importance of dynamic thresholding. Finally, we removed the feature extraction layer and
directly fed the images to the neuromodulated learning layer. In this case, we obtain an accuracy of
$90.72\%$, thus underscoring the importance of the feature extraction layer (sparse projection + dynamic 
thresholding in this work) in improving the accuracy.


\end{document}